\documentclass{article}

\usepackage{arxiv}
\usepackage{hyperref}
\hypersetup{%
colorlinks,
citecolor=blue,  
linkcolor=blue,  
urlcolor=black,  
hypertexnames=false,
hidelinks       
}
\usepackage{url}          
\usepackage{booktabs}      
\usepackage{nicefrac}       
\usepackage{microtype}      
\usepackage{lipsum}
\usepackage{graphicx}
\usepackage{caption}
\usepackage[section]{placeins}
\usepackage{subcaption}
\usepackage{float}
\usepackage[numbers,sort&compress]{natbib}
\usepackage{amsmath,amssymb,amsfonts}
\usepackage{algorithmic}
\usepackage{textcomp}
\usepackage{xcolor}

\graphicspath{ {./images/} }

\title{Active Label Cleaning for Reliable Detection of Electron Dense Deposits in Transmission Electron Microscopy Images}

\author{
  Jieyun Tan\thanks{These authors contributed equally.} \\
  School of Biomedical Engineering\\
  Southern Medical University\\
  Guangzhou, China \\
  \And
  Shuo Liu\footnotemark[1] \\
  School of Biomedical Engineering\\
  Southern Medical University\\
  Guangzhou, China \\
  \And
  Guibin Zhang\footnotemark[1] \\
  School of Biomedical Engineering\\
  Southern Medical University\\
  Guangzhou, China \\
  \And
  Ziqi Li \\
  College of Letters and Science\\
  University of Wisconsin-Madison\\
  Madison, USA \\
  \And
  Jian Geng \\
  School of Basic Medical Sciences\\
  Southern Medical University\\
  Guangzhou, China \\
  \And
  Lei Zhang \\
  Department of Nephrology\\
  Nanfang Hospital, Southern Medical University\\
  Guangzhou, China \\
  \And
  Lei Cao\thanks{Corresponding author: Lei Cao (\texttt{caolei@smu.edu.cn}).} \\
  School of Biomedical Engineering\\
  Southern Medical University\\
  Guangzhou, China \\
  \texttt{caolei@smu.edu.cn} \\
}

\begin{document}
\maketitle
\begin{abstract}
Electron dense deposits (EDD) provide crucial diagnostic references for glomerular disease. 
However, manual identification of EDD is labor-intensive and variable across observers, underscoring the need for automated and reliable image analysis tools. Yet, the accuracy of deep learning-based detection remains constrained by the lack of large, high-quality labeled datasets. Although crowdsourcing can lower annotation costs, it inevitably introduces label noise. 
To address this, we propose a novel active label cleaning method to effectively reduce label noise and improve data quality. 
We employ active learning to select the most valuable noisy samples for re-annotation by pathologists, constructing high-accuracy cleaning models at a minimal cost. 
These models are then utilized to perform a noise-graded correction on the crowdsourced labels. 
A core component, the Label Selection Module, leverages the inconsistency between crowdsourced labels and model predictions to achieve both image-level sample selection and instance-level noise grading. 
Experimental results show that our method achieves 67.18\% AP\textsubscript{50} on the private dataset, an 18.83\% improvement over noisy training. 
This performance reaches 95.79\% of that achieved with full pathologist annotation while reducing annotation cost by 73.30\%, 
ultimately delivering superior detection performance over existing noisy training methods at a lower cost.
These results demonstrate that the proposed active label cleaning method provides a practical and efficient solution for building reliable medical AI systems under limited expert annotation resources.
\end{abstract}

\keywords{Electron dense deposits, Object detection, Label cleaning, Active learning}

\section{Introduction}\label{sec1}

Electron dense deposits (EDD) are immune complexes visible in transmission electron microscopy (TEM) images which can 
provide crucial diagnostic references for glomerular disease \cite{b1,b22}. 
Recent progress in deep learning leads to substantially improvements in automatic detection of EDD \cite{b2,b3}, 
which has the potential to facilitate pathologists' diagnostic work. Nevertheless, due to the scarcity of labeled large-scale datasets, 
the model's detection accuracy does not achieve the requisite for clinical application. 
There are two main factors contributing to EDD annotation scarcity. 
Firstly, medical image annotation requires professional expertise and typically involves the collaboration of three or more experts to ensure the reliability \cite{b4}. 
Secondly, the annotation process for EDD in TEM images is costly and time-consuming because of their large quantity, 
diverse morphology, and similarity to other ultra-structures in terms of shape and grayscale distribution.

In the medical field, crowdsourcing (data labeling through non-expert annotators) is an effective method to addresses insufficient labeling \cite{b5}, 
but it inevitably introduces noise \cite{b6}. Hence, research on effective methods to reduce the influences of noise within datasets is crucial \cite{b7}. 
Among existing methods, noise-robust learning \cite{b8,b9} improves efficiency by directly training models on noisy data without label correction. 
However, it requires substantial training data to cover various types and levels of noises to make up accuracy loss, 
which is inappropriate for data-limited clinical scenarios. 
In contrast, label cleaning selects samples (i.e., images) and corrects noisy labels by establishing a cleaning model \cite{b10,b11}, 
which theoretically has higher detection accuracy. 
However, if the cleaning model is trained solely on noisy data, its training signals (e.g., loss, gradient) cannot reliably reflect true noise level, 
leading to lower accuracy in sample selection and label correction. 
Moreover, constrained by the training algorithm and data, exclusively relying on automatically correction of noisy labels via the cleaning model presents a latent risk of learning the bias in the noise, 
which may further amplify label noise \cite{b20}.

To mitigate label noise and improve data quality, 
(1) we designed a two-step process to construct high-accuracy cleaning models that perform label correction using only datasets labeled by pathologists. 
Clean label acquisition is costly, so (2) we introduce an active learning method to perform image-level sample selection in the first step, maximizing the effectiveness of training data while controlling annotation cost. 
We use the proposed Label Selection Module (LSM) to analyze the inconsistency between crowdsourced labels and model predictions, which helps us to select the most valuable samples for labeling. 
In the second step, (3) we perform instance-level label correction (where each instance corresponds to one label) based on the LSM's discrimination results: 
labels with simple noise are automatically corrected by the model, while those with complex noise are assigned to pathologists for review. 
This approach significantly reduces annotation cost while improving the reliability of the corrections.

The remainder of this paper is structured as follows: Section \ref{sec2} introduces related works on object detection with noisy label and active learning; Section \ref{sec3} presents a detailed description of the private dataset and the proposed method; Section \ref{sec4} reports the experimental settings and results; and Section \ref{sec5} concludes the paper.

\section{Related work}\label{sec2}

\subsection{Object detection with noisy label}
In object detection tasks with noisy label, the smallest unit to measure noise level is the bounding box(i.e., instance or label). 
Thus, the noise types include not only classification error, but also background noise, miss noise, location noise, etc. \cite{b13}, as illustrated in Figure~\ref{fig1}. 
In current research, some methods aim to improve the robustness of the model on data with noisy label. 
For example, NOTE-RCNN \cite{b8} avoids involving complex noisy labels in training via semi-supervised methods, 
while OA-MIL \cite{b9} designs robust training strategies for incomplete and imprecise bounding boxes. 
However, these methods only enhance the model's noise tolerance at the algorithmic level but are unable to eliminate the noise, and the residual label noise still limits the achievable accuracy, especially in high-precision tasks such as medical image analysis.

Other approaches leverage models trained on noisy data for sample selection and label correction, thereby eliminating label noise. 
For example, CA-BBC \cite{b10} corrects location errors in noisy labels using gradient information and adjusts classification errors by model predictions, yet updates all labels indiscriminately. 
Mao et al. \cite{b11} re-annotate those noisy labels which significantly differ from previous predictions in each epoch. 
Although such selection-based corrections can reduce obvious errors, methods that depend primarily on a single model's predictions remain vulnerable when model reliability degrades under high noise. 
This dependency limits the extent to which automatic correction can approach expert-level annotations in noisy scenarios.

\subsection{Active learning}
Active learning aims to minimize annotation cost by selecting the most informative training samples for improving model accuracy \cite{b12,b14}. 
In object detection tasks, representative works include LT/C and LS+C \cite{b15}, which assess image information for sample selection by calculating the tightness and stability of instances, respectively. 
Meanwhile, CALD \cite{b16} selects samples by analyzing the inconsistency of predictions before and after data augmentation. 
More recently, OSAL-ND \cite{b17} first selects candidate samples through prototypes and then queries the most uncertain samples for re-annotation. 
For task-unaware active learning methods, VAAL \cite{b18} and LLAL \cite{b19} leverage image features or training signals to select samples by calculating the latent features or losses.

In addition, active label cleaning (ALC) has attracted wide attention as an essential application of active learning for data with noisy label, particularly for medical image processing \cite{b20}. 
For example, Khanal et al. \cite{b21} propose a two-phase approach combining noisy label learning and active label cleaning to improve the robustness of medical image classification under limited budgets. 
However, to our knowledge, ALC methods that explicitly target instance-level label noise and the spatial complexity of detection remain relatively underexplored.

\section{Methods}\label{sec3}
\subsection{Dataset}
\begin{figure}[htbp]
\includegraphics[width=1.0\linewidth]{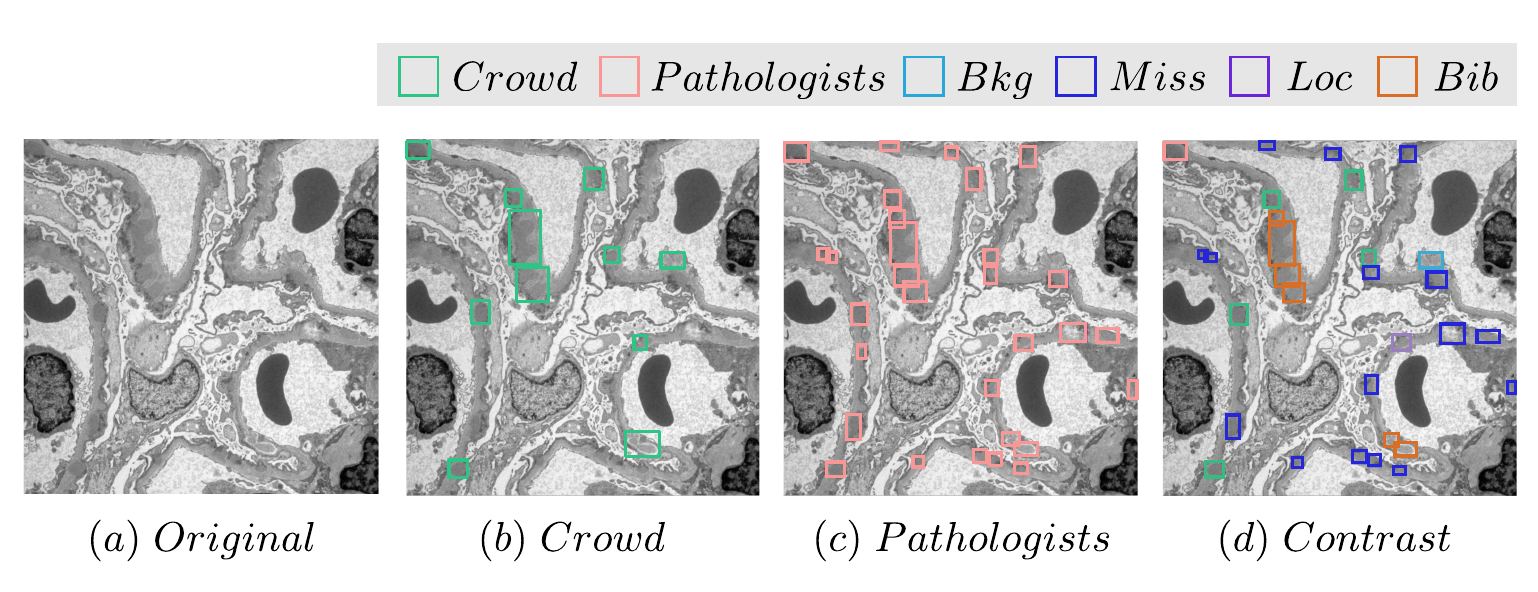}
\caption{Classification of labels. (a) Original TEM images; (b) Crowdsourced labels with noise (green bounding boxes); (c) Pathologists-corrected clean labels (pink bounding boxes); (d) Comparison of annotation differences, categorized into: Bkg (background noise, background misclassified as foreground, light blue), Miss (miss noise, ground truth missed by annotators, deep blue), Loc (location noise, boxes incorrectly localized, purple), and Bib (box-in-box noise, boxes incorrectly nested, orange).}
\label{fig1}
\end{figure}

\begin{figure}[htbp]
\includegraphics[width=0.85\linewidth]{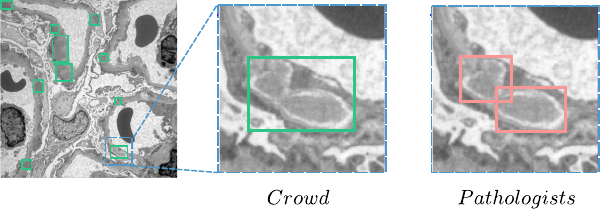}
\caption{Bib (Box-in-Box) Noise in crowdsourced labels. Blurred EDD edges in the dataset led crowd annotators to occasionally include multiple EDD within a single bounding box. While pathologists avoid this error, it reflects a common localization artifact in crowdsourced label.}
\label{fig2}
\end{figure}

This study was a retrospective study. The private dataset was collected from a certified medical laboratory center, 
all data has been anonymized, personal information excluded. 
A total of 1112 glomerular TEM images (2,048$\times$2,048) were acquired from 202 patients with membranous nephropathy who underwent renal biopsy, 
each image contained one or more EDD: 131 images were randomly selected for validation, 234 for testing, and the rest for training. 
The annotators include 5 medical students with basic training, serving as crowd, and 3 senior pathologists with over a decade of experience. 
Crowd-annotated labels may contain noise, whereas pathologist-annotated labels are considered clean. 
We conducted a comparative analysis of these two types of labels to effectively clean the noise data, as shown in Figure~\ref{fig1}. 
Meanwhile, Figure~\ref{fig2} shows an example of Bib noise, which occurs when blurred EDD edges cause crowd annotators to box multiple instances together.

\subsection{Framework Overview}
\begin{figure}[htbp!]
\includegraphics[width=1.0\linewidth]{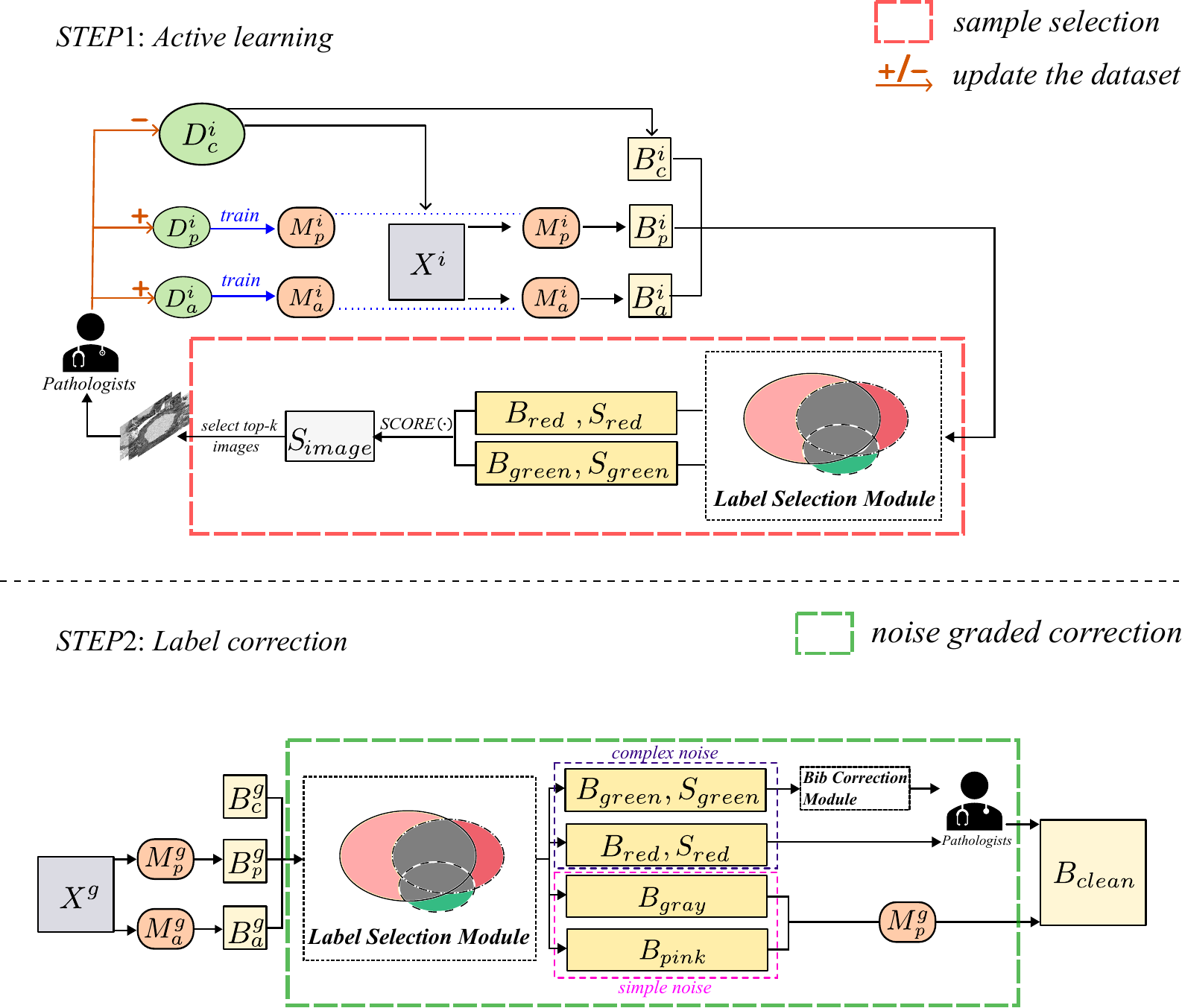}
\caption{Overview of the proposed method.}
\label{fig3}
\end{figure}

The proposed method involves two steps: Step 1 is Active learning, which is used to train cleaning models. 
Step 2 is Label correction, which corrects noisy labels using the cleaning models constructed in Step 1. 
Both steps rely on the Label Selection Module. The entire process is illustrated in Figure~\ref{fig3}.

\subsection{Initialization}
As illustrates in Figure~\ref{fig3}, $X$ represents the entire set of images. 
Initially ($i=0$), $X^0$ (a subset of images) is randomly selected from $X$ and provided to pathologists for annotation, 
resulting in the initial clean dataset $D_p^0$. The corresponding crowdsourced dataset is $D_c^0$. 
Next, labels corresponding to EDD that are validated by both crowd and pathologists are selected to construct the consensus dataset $D_a^0$.
After obtaining these initial datasets, two architecturally identical models are initialized. Based on their respective training datasets, they are denoted as $M_p^0$ and $M_a^0$.
The former is a cleaning model trained on the clean dataset, while the latter is trained on the consensus dataset.

\subsection{Label Selection Module}
\begin{figure}[htbp]
\includegraphics[width=0.95\linewidth]{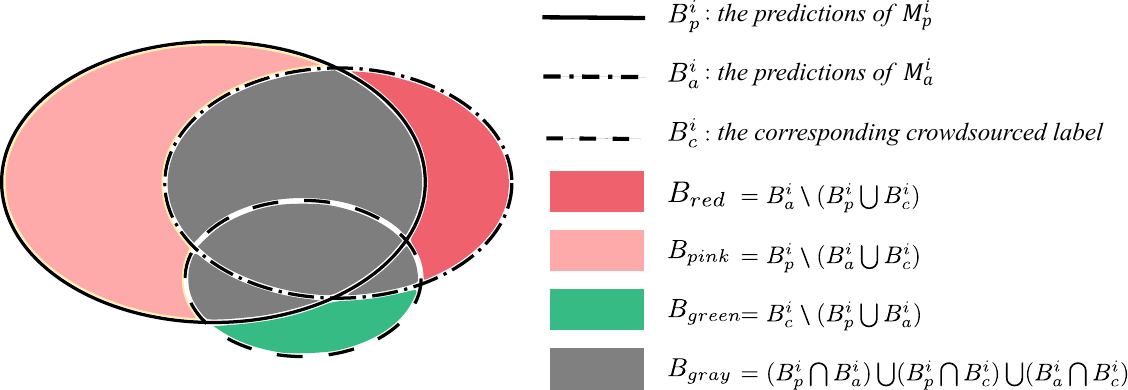}
\caption{Label Selection Module diagram.}
\label{fig4}
\end{figure}

As illustrated in Figure~\ref{fig4}, the primary function of the Label Selection Module (LSM) in the $i$-th iteration is to identify potential noisy labels by comparing crowdsourced labels with the predictions of two models ($M_p^i$ and $M_a^i$). 
It then classifies these labels based on their degree of inconsistency and assigns each a quantified score, thereby providing a basis for following active learning and label cleaning.
Specifically, $B_a^i$ represents the prediction results of $M_a^i$. 
Due to the incompleteness of training data, it may contain Miss noise. 
In contrast, $B_p^i$ represents the prediction results of $M_p^i$ trained using all available clean labels. 
While $M_p^i$ can detect more EDD, it is more sensitive to abnormal cases because its training data includes more anomalous instances, particularly atypical EDD. 
This heightened sensitivity, especially pronounced in early stages with limited training data where the model may overfit, can lead to a higher propensity for false positives (Bkg noise) in its predictions, though this effect attenuates as training data increases.

In this module, we categorize label regions as follows:

\begin{itemize}
    \item \textbf{Red Region} ($B_{\text{red}}$): This region indicates labels in the detection results $M_a^i$ that do not consent with the results from $M_p^i$ or the crowdsourced labels $B_c^i$. These labels show significant inconsistency and usually contain substantial noise.

    \item \textbf{Pink Region} ($B_{\text{pink}}$): This region represents labels that only appear in the detection results of $M_p^i$, potentially due to omissions by crowd. As $M_p^i$ is trained on high-quality data, it can reliably identify these missing EDD annotations, thereby correcting the Miss noise contained within this region.

    \item \textbf{Green Region} ($B_{\text{green}}$): This region represents labels that only appear in the crowdsourced labels $B_c^i$, probably incorrectly annotated by crowd annotators. These labels are often associated with samples that are challenging to identify, may contain complex noise including Bkg, Loc, Bib noise.

    \item \textbf{Grey Region} ($B_{\text{gray}}$): This region indicates labels with high consistency between $B_c^i$, $B_p^i$, and $B_a^i$, specifically when the Intersection over Union (IoU) of instances in any two sets is greater than 0.5. These labels correspond to EDD that the model can accurately identify, typically containing no noise or only a small amount of noise.
\end{itemize}

We prioritize complex noisy labels ($B_{\text{red}}$ and $B_{\text{green}}$) over the simple noisy labels ($B_{\text{gray}}$ and $B_{\text{pink}}$) that can be automatically corrected. 
To select the most valuable images for active learning, we need to assess the inconsistency of labels within each image. 
Specifically, for each label in $B_{\text{red}}$, we use the confidence score from $M_a^i$'s prediction as the evaluation score, yielding the score set $S_{\text{red}}$ corresponding to $B_{\text{red}}$. 
For each label in $B_{\text{green}}$, we use the confidence score of the bounding box with the highest IoU from $M_p^i$'s prediction as the evaluation score, yielding the score set $S_{\text{green}}$ corresponding to $B_{\text{green}}$.

\subsection{Step 1: Active learning}
Figure~\ref{fig3} reveals that Step 1 iteratively select and annotate samples using active learning to construct the cleaning models. 
In the $i$-th iteration, the consensus dataset $\mathit{D}_a^i$ is built using the clean dataset $\mathit{D}_p^i$ and the corresponding crowdsourced dataset $\mathit{D}_c^i$.
Let $\mathit{X}^i = \mathit{X}^{i-1} \cup \mathit{A}^i$ (where $i \geq 1$) denote the current accumulated image set, with $\mathit{X}^0$ as the initial set and $\mathit{A}^i$ representing newly added images in the $i$-th iteration. 
For each image in $\mathit{X}^i$, if there exists a label $b_p$ from $\mathit{D}_p^i$ and a label $b_c$ from $\mathit{D}_c^i$ satisfy Equation~\eqref{eq1} (where $\delta$ is the threshold), 
the consensus labels $\mathit{B}_p$ (all matching $b_p$ instances) are added to $\mathit{D}_a^{i-1}$ to form $\mathit{D}_a^i$.


\begin{equation}
\begin{aligned}
&\mathit{D}_a^i \leftarrow \mathit{D}_a^{i-1} \cup \big\{ (x, \mathit{B}_p) \mid x \in \mathit{X}^i, \mathit{B}_p = \{ b_p \in \mathit{D}_p^i \mid \exists b_c \in \mathit{D}_c^i: b_c \cap b_p/b_p > \delta \} \big\}
\end{aligned}
\label{eq1}
\end{equation}

Subsequently, $M_p^i$ and $M_a^i$, are trained using $\mathit{D}_p^i$ and $\mathit{D}_a^i$, respectively. 
They detect EDD in the remaining image set $\mathit{X} \setminus \mathit{X}^i$, producing results $\mathit{B}_p^i$ and $\mathit{B}_a^i$. 
The LSM then analyzes the inconsistencies among $\mathit{B}_p^i$, $\mathit{B}_a^i$ and the corresponding crowdsourced labels $\mathit{B}_c^i$, 
selecting complex noisy label sets $\mathit{B}_{\text{red}}$ and $\mathit{B}_{\text{green}}$, along with their evaluation score sets $\mathit{S}_{\text{red}}$ and $\mathit{S}_{\text{green}}$. 
The image scores are then aggregated to select samples. 
For the $j$-th image $x_j \in \mathit{X} \setminus \mathit{X}^i$, an aggregation function $\mathit{SCORE}(\cdot)$ computes its inconsistency score by summing instance-level evaluation scores from $\mathit{S}_{\text{red}}$ and $\mathit{S}_{\text{green}}$ associated with labels within $x_j$.

\begin{equation}
\mathit{SCORE}(x_j) = \sum_{l \in \mathit{B}_{\text{red}} \cup \mathit{B}_{\text{green}}} \left( s_{\text{red}}^l + s_{\text{green}}^l \right)
\label{eq2}
\end{equation}

\noindent
where $s_{\text{red}}^l$ and $s_{\text{green}}^l$ denote the instance-level scores for label $l$ in $\mathit{S}_{\text{red}}$ and $\mathit{S}_{\text{green}}$, respectively.

After quantifying label inconsistency, the top-$k$ most inconsistent images (indicating highly noisy labels or difficult-to-identify EDD) are selected for pathologist review. 
Corrected labels are then used to expand $\mathit{D}_p^i$, generating the updated clean dataset $\mathit{D}_p^{i+1}$. 
Simultaneously, the corresponding noisy labels are removed from $\mathit{D}_c^i$ to produce $\mathit{D}_c^{i+1}$, and $\mathit{D}_a^{i+1}$ is expanded following Equation~\eqref{eq1}. 
This process iterates until the stopping condition is met at the $g$-th iteration, yielding final cleaning models $M_p^g$ and $M_a^g$ for label correction.

\subsection{Step 2: Label correction}
Figure~\ref{fig3} shows that Step 2 performs instance-level noise-graded correction on the remaining noisy labels in $\mathit{D}_c^g$, leveraging the cleaning models $\mathit{M}_p^g$ and $\mathit{M}_a^g$ constructed in Step 1. 
First, we apply $\mathit{M}_p^g$ and $\mathit{M}_a^g$ to detect EDD in the remaining image set $\mathit{X} \setminus \mathit{X}^g$, producing results $\mathit{B}_p^g$ and $\mathit{B}_a^g$. 
These results, along with the corresponding noisy labels $\mathit{B}_c^g$, are fed into the LSM, which classifies the labels into four categories: $\mathit{B}_{\mathrm{gray}}$, $\mathit{B}_{\mathrm{pink}}$, $\mathit{B}_{\mathrm{red}}$ and $\mathit{B}_{\mathrm{green}}$.

For labels with simple noise ($\mathit{B}_{\mathrm{gray}}$ and $\mathit{B}_{\mathrm{pink}}$), we automatically correct them by adding missing labels or replacing incorrect ones in $\mathit{D}_c^g$ with the predictions $\mathit{B}_p^g$ from $\mathit{M}_p^g$, as $\mathit{M}_p^g$ was trained on the fully clean dataset and demonstrates superior detection performance to $\mathit{M}_a^g$. 
Labels with complex noise ($\mathit{B}_{\mathrm{red}}$ and $\mathit{B}_{\mathrm{green}}$) are submitted to pathologists for manual correction, with $\mathit{B}_p^g$ and $\mathit{B}_a^g$ provided as reference suggestions.

Notably, the LSM is unable to identify Bib noise in some labels within $\mathit{B}_{\mathrm{green}}$. 
Since such noise can be auto-corrected to reduce pathologists' time. Inspired by \cite{b23}, 
we developed a Bib Correction Module before manual review. For each label $b_{\mathrm{green}} \in \mathit{B}_{\mathrm{green}}$, 
if there exists $b^\ast$ (where $b^\ast \in \mathit{B}_{\mathrm{gray}} \cup \mathit{B}_{\mathrm{pink}} \cup \mathit{B}_{\mathrm{red}}$) satisfying Equation~\eqref{eq3} (where $\gamma$ is the threshold), 
$b_{\mathrm{green}}$ is identified as Bib noise and removed, while $b^\ast$ is retained as the correct label.

\begin{equation}
\begin{split}
& \mathit{B}_{\mathrm{green}} \leftarrow \mathit{B}_{\mathrm{green}} \setminus  \Bigl\{ b_{\mathrm{green}} \Bigm| \exists b^* \in \mathit{B}_p^g \vee b^* \in \mathit{B}_a^g : 
\frac{b^* \cap b_{\mathrm{green}}}{b^*} > \gamma \Bigr\}
\end{split}
\label{eq3}
\end{equation}

The final clean labels $\mathit{B}_{\mathrm{clean}}$ are generated by combining auto-corrected and manually corrected labels, thereby completing the noise-graded correction for remaining noisy labels in $\mathit{D}_c^g$.

\section{Experimental results}\label{sec4}
\subsection{Experimental settings}
All experiments were conducted on an RTX 3090 GPU using \textsc{MMDetection} \cite{b24}. We employed \textsc{Faster R-CNN} \cite{b25} with a \textsc{ResNet-50} (pre-trained on ImageNet) + \textsc{FPN} backbone, trained for 36 epochs with batch size of 4. The \textsc{SGD} optimizer was configured with a learning rate of 0.01, momentum of 0.9, and weight decay of $1\times10^{-4}$. To enhance model robustness, data augmentation was applied using random flipping, rotation, scaling, and clipping.
In Step 1, active learning began with 40 randomly selected images ($|\mathit{X}^0| = 40$). The consensus dataset $\mathit{D}_a^i$ was constructed using a threshold $\delta = 0.5$, with 4 iterations ($g = 4$) selecting 40 samples ($k = 40$) per iteration, culminating in 200 samples (26.77\% of total) for final models training. 
In Step 2, the Bib Correction Module's threshold $\gamma$ was set to 0.8. We evaluated model performance using $\mathrm{AP}_{50}$ and assessed label cleaning effectiveness through \textsc{TIDE} \cite{b26} analysis of three noise types. The annotation cost ratio between crowd and pathologist efforts was maintained at $1:10$. All experiments were repeated three times with averaged results reported.

\subsection{Comparison Results}

Table~\ref{tab1} shows the evaluation results of our proposed method on the private dataset, which was retrained $M_p$ using $D_p^g$ after Step 2, and evaluated on the test set. 
Compared with other methods, our approach achieved the best performance with an $\mathrm{AP}_{50}$ of 67.18\%, demonstrating an 18.83\% improvement over the uncleaned baseline of Noisy Training and reaching 95.79\% of the performance of the Clean Training (a model trained on pathologist-annotated data). 
This indicates that our method enables robust training in practical applications. 
Furthermore, TIDE analysis reveals that our approach significantly reduces all three types of noise. 

\begin{table*}[htbp]
	\centering
	\caption{Comparison of different noise treatment methods on private dataset.}
	\label{tab1}
	\renewcommand{\arraystretch}{1.3} 
	\begin{tabular*}{\textwidth}{@{}@{\extracolsep{\fill}}lcccccc@{}}
	\toprule[1.5pt]
	Method &
	$\mathrm{AP}_{50}(\%)$$\uparrow$ & 
	\multicolumn{3}{c}{TIDE Analysis} & 
	Budget(\%)$\downarrow$ \\
	\cmidrule{3-5}
	 & & Bkg$\downarrow$ & Miss$\downarrow$ & Loc$\downarrow$ & \\
	\midrule
	Noisy Training & 48.35 & 5.34 & 15.94 & 17.05 & 3.52 \\
	\midrule
	OA-MIL & 49.90{(+1.55)} & 3.99{(-1.35)} & 14.77{(-1.17)} & 20.90{(+3.85)} & \textbf{3.52} \\
	Mao & 56.23{(+7.88)} & 4.92{(-0.42)} & 16.34{(+0.40)} & 13.63{(-3.42)} & \textbf{3.52} \\
	ALC & 63.67{(+15.32)} & 4.32{(-1.02)} & 11.77{(-4.17)} & 12.65{(-4.40)} & 26.71{(+23.19)} \\
	Ours (Random) & 64.07{(+15.72)} & 4.33{(-1.01)} & 12.06{(-3.88)} & 11.70{(-5.35)} & 31.68{(+28.16)} \\
	Ours & \textbf{67.18{(+18.83)}} & \textbf{3.87{(-1.47)}} & \textbf{10.70{(-5.24)}} & \textbf{11.30{(-5.75)}} & 26.70{(+23.18)} \\
	\midrule
	Clean Training & 70.13{(+21.78)} & 2.83{(-2.51)} & 10.94{(-5.00)} & 11.07{(-5.98)} & 100.00{(+96.48)} \\
	\bottomrule[1.5pt]
	\end{tabular*}%
	
	\vspace{4pt}
	\raggedright
	\footnotesize
	\textbf{Note:} Values in parentheses denote changes relative to Noisy Training. 
	$\uparrow$/$\downarrow$ indicate that higher/lower values are better, respectively. 
	\textbf{Bold} denotes the best performance in each column.
	\textbf{Budget} column denotes the sum of the crowd's costs and correction costs and the pathologist's costs. 
	\textbf{Noisy Training} serves as the baseline, where the model is trained entirely on the uncleaned crowdsourced dataset $D_c$. 
	\textbf{Ours (Random)} replaces the sample selection strategy in Step 1 with random selection, aiming to validate the effectiveness of the active learning approach. 
	\textbf{Clean Training} refers to models fully supervised trained on the clean dataset $D_p$, which entirely annotated by pathologists.
\end{table*}

In contrast, OA-MIL, which focuses on noise-robust learning without cleaning labels at the source, exhibits limited noise reduction efficacy and even increases Loc noise. 
Among label cleaning methods, Mao trained a cleaning model using noisy labels, but its performance was compromised by data noise, leading to inferior noise correction. 
In comparison, ALC, by incorporating active learning, it enhancing label cleaning effectiveness through pathologist involvement, outperforming Mao's purely model-driven automatic correction, particularly for complex noise. 
Additionally, ALC's active learning framework was not specifically designed for object detection tasks, limiting its performance, as also observed in Ours (Random). 
In summary, our proposed method achieves superior label correction quality while significantly reducing annotation cost---demonstrating a 73.30\% reduction compared to Clean Training.

\subsection{Qualitative Analysis of Noise Improvement}

\begin{figure}[htbp]
	\centerline{\includegraphics[width=0.8\linewidth]{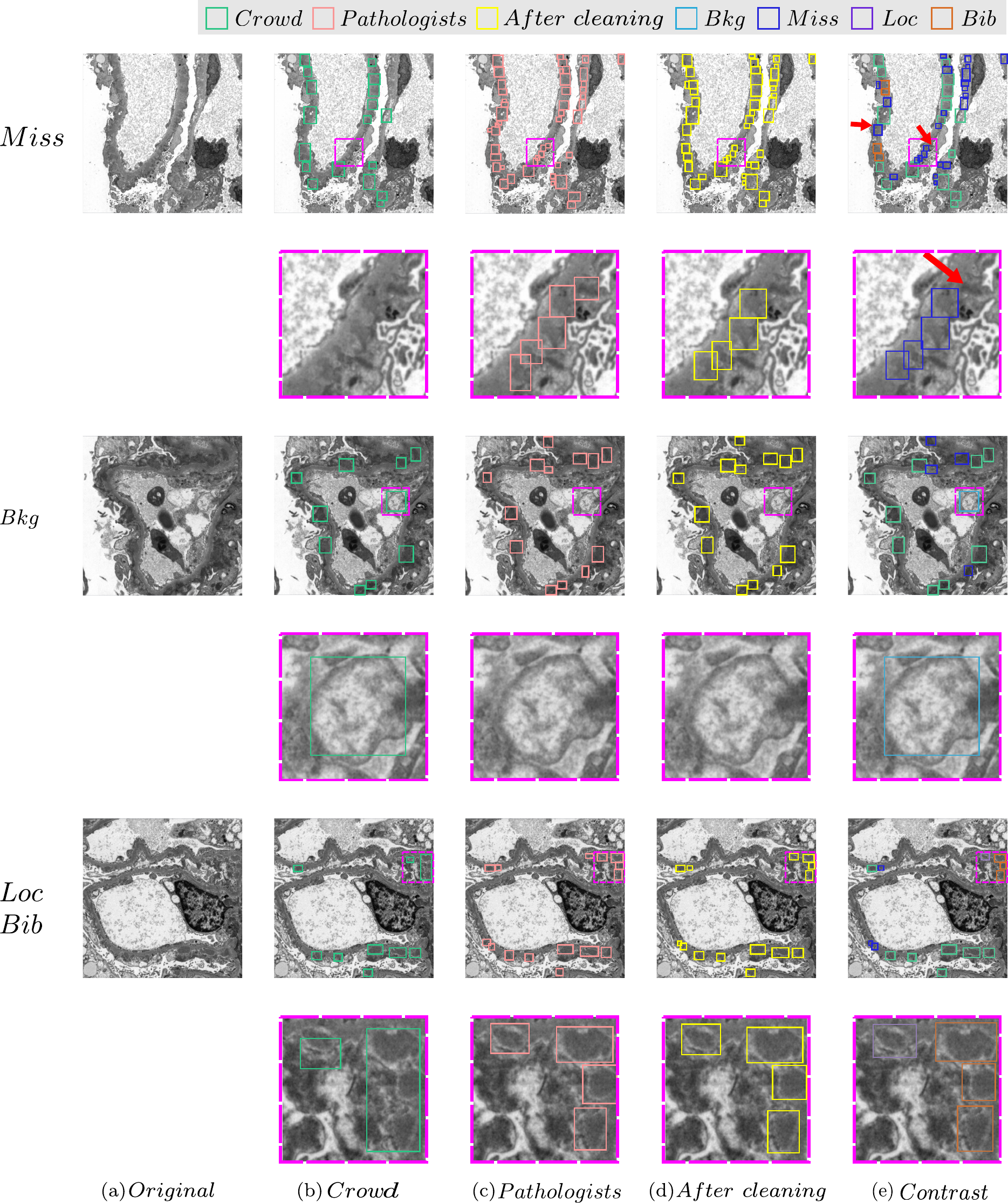}}
	\caption{Qualitative results of label noise correction.}
	\label{fig5}
\end{figure}

To visually assess the effectiveness of our proposed noise correction method, qualitative results are presented in Figure~\ref{fig5}. 
The figure illustrates representative examples of various noise types before and after label correction. 
As shown in the first and second rows, our method successfully recovered several instances that were previously missed by the crowd (Miss). 
However, not all missing instances could be fully corrected, as indicated by the red arrow in Figure~\ref{fig5}(e), which highlights a few labels still absent compared to pathologist annotations. 
In addition, the third and fourth rows demonstrate that our method effectively removed false positive labels on background regions (Bkg). 
Furthermore, the fifth and sixth rows show that the proposed method identified and corrected the bounding boxes that were either misaligned or contained other boundaries (Loc and Bib). 
These visual results validate that our method can substantially improve annotation quality across multiple types of label noise.

\FloatBarrier

\subsection{Generalizability Study}
To validate the generalization ability of the proposed method, we conducted experiments on the public BCCD dataset~\cite{b28}, comprising 205 training images, 87 validation images, and 72 test images. 
We simulated crowdsourced label by injecting Bkg, Miss, Loc and Bib noise into the training data at the same ratios as in the private dataset. 
As shown in Table~\ref{tab2}, our method consistently achieved the best performance on the BCCD dataset, demonstrating an $\mathrm{AP}_{50}$ of 88.63\%, a 4.81\% improvement over the uncleaned baseline of Noisy Training, while significantly reducing all three noise types and reaching 97.97\% of the Clean Training performance. 
Notably, the results revealed that OA-MIL outperformed Mao and ALC surpassed Ours(Random), which contrasts with the findings in Table~\ref{tab1}. 
We attribute this difference to the BCCD dataset's limited scale, Mao's dependence on the amount of data, and the inherent high randomness in Ours (Random).

\vspace{-6pt}

\begin{table*}[htbp]
	\centering
	\caption{Comparison of different noise treatment methods on public dataset.}
	\label{tab2}
	\renewcommand{\arraystretch}{1.3} 
	\begin{tabular*}{\textwidth}{@{}@{\extracolsep{\fill}}lcccccc@{}}
	\toprule[1.5pt]
	Method &
	$\mathrm{AP}_{50}(\%)$$\uparrow$ & 
	\multicolumn{3}{c}{TIDE Analysis} & 
	Budget(\%)$\downarrow$ \\
	\cmidrule{3-5}
	 & & Bkg$\downarrow$ & Miss$\downarrow$ & Loc$\downarrow$ & \\
	\midrule
	Noisy Training & 83.82 & 10.15 & 0.77 & 2.42 & 3.43 \\
	\midrule
	OA-MIL & 84.56{(+0.74)} & \textbf{5.25{(-4.90)}} & 0.79{(+0.02)} & 4.67{(+2.25)} & \textbf{3.43} \\
	Mao & 84.01{(+0.19)} & 8.53{(-1.62)} & 0.83{(+0.06)} & 2.64{(+0.22)} & \textbf{3.43} \\
	ALC & 87.07{(+3.25)} & 7.37{(-2.78)} & 0.63{(-0.14)} & 1.07{(-1.35)} & 29.87{(+26.44)} \\
	Ours (Random) & 86.44{(+2.62)} & 7.61{(-2.54)} & 0.59{(-0.18)} & 1.09{(-1.33)} & 30.26{(+26.83)} \\
	Ours & \textbf{88.63{(+4.81)}} & 7.26{(-2.89)} & \textbf{0.57{(-0.20)}} & \textbf{0.69{(-1.73)}} & 29.91{(+26.48)} \\
	\midrule
	Clean Training & 90.47{(+6.65)} & 6.68{(-3.47)} & 0.32{(-0.45)} & 0.69{(-1.73)} & 100.00{(+96.57)} \\
	\bottomrule[1.5pt]
	\end{tabular*}%
	
	\vspace{4pt}
	\raggedright
	\footnotesize
	\textbf{Note:} Values in parentheses denote changes relative to Noisy Training. 
	$\uparrow$/$\downarrow$ indicate that higher/lower values are better, respectively. 
	\textbf{Bold} denotes the best performance in each column.
	\textbf{Budget} column denotes the sum of the crowd's costs and correction costs and the pathologist's costs.
\end{table*}

\vspace{-12pt}

\subsection{Ablation Results}
\begin{figure}[htbp]
\centerline{\includegraphics[width=0.9\linewidth]{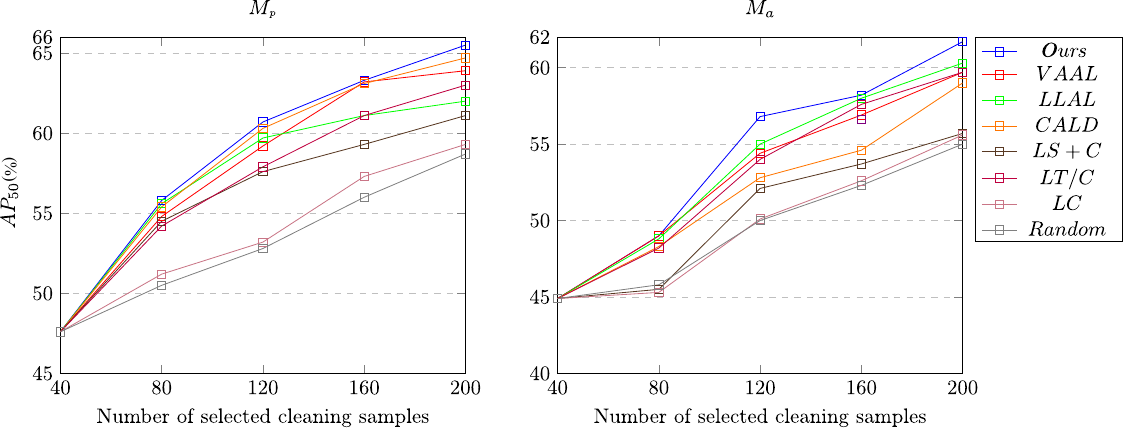}}
\caption{Influence of different Active Learning methods.}
\label{fig6}
\end{figure}

\textbf{The influence of active learning methods:} To evaluate the impact of different active learning methods on the performance of the two cleaning models, $\mathit{M_p}$ and $\mathit{M_a}$, we compared our proposed approach with random selection (Random), task-unaware methods (VAAL and LLAL), classification-specific method (LC~\cite{b27}), and detection-specific methods (LS+C, LT/C and CALD). 
As shown in Figure~\ref{fig6}, all active learning methods outperformed Random, with the classification-specific method LC exhibiting inferior performance due to task mismatch. Moreover, detection-specific methods (LS+C, LT/C, CALD) and task-unaware approaches (VAAL, LLAL) achieved comparable results, likely because they were primarily designed for large-scale natural images and are less suitable for our small-scale dataset. 
In contrast, our method is specifically designed for EDD detection and demonstrates superior performance under limited data conditions. When the training samples increased to 200, our approach significantly surpassed all other methods in model accuracy, improving $\mathrm{AP}_{50}$ by 6.80\% for $\mathit{M_p}$ and 6.70\% for $\mathit{M_a}$ compared to the Random.

\FloatBarrier

\textbf{The influence of Bib Correction Module:} We evaluated the impact of the Bib Correction Module on both label correction effectiveness and the time required for pathologists' correction, while conducting an ablation study on the Bib noise threshold $\gamma$ in Equation~\eqref{eq3}. As shown in Table~\ref{tab3}, the introduction of the module successfully reduced the time pathologists spent on label corrections by about 10\% without affecting the label correction effectiveness. Moreover, this improvement remained consistent across different threshold values of $\gamma$.

\begin{table}[htbp]
	\centering
  \caption{Evaluating the impact of the Bib Correction Module on label correction time.}
	\label{tab3}
	\renewcommand{\arraystretch}{1.3} 
	\begin{tabular*}{\textwidth}{@{}@{\extracolsep{\fill}}lccc@{}}
	\toprule[1.5pt]
	Bib module & $\gamma$ & $\mathrm{AP}_{50}(\%)$$\uparrow$ & Time(\%)$\downarrow$ \\
	\midrule
	$\times$ & -- & 67.17 & 100.00 \\
	$\surd$ & 0.7 & \textbf{67.18} & \textbf{90.37} \\
	$\surd$ & 0.8 & \textbf{67.18} & 90.96 \\
	$\surd$ & 0.9 & 67.17 & 92.77 \\
	\bottomrule[1.5pt]
	\end{tabular*}
	
	\vspace{4pt}
	\raggedright
	\footnotesize
	\textbf{Note:} 
	$\uparrow$/$\downarrow$ indicate that higher/lower values are better, respectively. 
	\textbf{Bold} denotes the best performance in each column.
	\textbf{Time} is the ratio of pathologist's label correction time when using the Bib Correction Module versus without it.
\end{table}

\textbf{The influence of $\mathit{M_a}$ on sample selection and label correction:} 
To evaluate the effect of $\mathit{M_a}$, we compared the performance of the cleaning model $\mathit{M_p}$ with and without the participation of $\mathit{M_a}$. As shown in Table~\ref{tab4}, incorporating $\mathit{M_a}$ improves the detection accuracy of $\mathit{M_p}$ by 2.17\%, enhancing the efficiency of sample selection. During the label correction process, introducing $\mathit{M_a}$ further improves accuracy by 3.88\%. Moreover, combining the uncertainty of $\mathit{M_p}$ and $\mathit{M_a}$ significantly reduces Loc by 7.38\% in the TIDE analysis, demonstrating the superior effectiveness of the dual-model collaboration.

\begin{table}[htbp]
	\centering
	\caption{Evaluating the impact of the $\mathit{M_a}$ model on sample selection and label correction.}
	\label{tab4}
	\renewcommand{\arraystretch}{1.3} 
	\begin{tabular*}{\textwidth}{@{}@{\extracolsep{\fill}}lccccc@{}}
	\toprule[1.5pt]
	$\mathit{M_a}$ & \multicolumn{2}{c}{$\mathrm{AP}_{50}(\%)$$\uparrow$} & \multicolumn{3}{c}{TIDE Analysis} \\
	\cmidrule(lr){2-3} \cmidrule(lr){4-6}
	 & Cleaning Model & After Cleaning & Bkg$\downarrow$ & Miss$\downarrow$ & Loc$\downarrow$ \\
	\midrule
	$\times$ & 63.33 & 63.30 & 22.28 & 12.85 & 19.13 \\
	$\surd$ & \textbf{65.50}(+2.17) & \textbf{67.18}(+3.88) & 22.16(-0.12) & 12.65(-0.20) & \textbf{11.75}(-7.38) \\
	\bottomrule[1.5pt]
	\end{tabular*}
	
	\vspace{4pt}
	\raggedright
	\footnotesize
	\textbf{Note:} 
	Values in parentheses denote changes relative to the baseline ($\mathit{M_a}$=$\times$).
	$\uparrow$/$\downarrow$indicate that higher/lower values are better, respectively. 
	\textbf{Bold} denotes the best performance in each column.
\end{table}

\section{Conclusions}\label{sec5}

We propose an active label cleaning method for addressing label noise in crowdsourced label for EDD detection, which balances performance and cost. 
Specifically, we employ active learning to select the most valuable samples, constructing the high-accuracy cleaning models with minimal pathologist's annotation. 
Subsequently, we implement noise-graded correction to reduce annotation cost without compromising label correction quality. 
The designed Label Selection Module analyzes the inconsistency between the crowdsourced labels and the model predictions to achieve image-level sample selection and instance-level noise grading, serving as an effective auxiliary tool. 
Finally, experimental results demonstrate that the proposed method achieves optimal results on both private and public datasets, confirming its effectiveness and generalizability.

\section*{Author contributions}

Conceptualization, J.T.; methodology, J.T. and S.L.; software, J.T. and S.L.; validation, J.T. and S.L.; data curation, J.G.; writing---original draft preparation, J.T.; writing---review and editing, J.T., S.L., G.Z, Z.L., L.Z and L.C.; supervision, L.C.; funding acquisition, L.Z. and L.C. All authors have read and agreed to the published version of the manuscript.

\section*{Acknowledgments}
This research was funded by the grant from the National Natural Science Foundation of China (No. 32071368), the Guangdong Basic and Applied Basic Research Foundation (No. 2022A1515110162) and the Clinical Research Fund of Nanfang Hospital, Southern Medical University (No. 2023CR024).

\section*{Financial disclosure}

None reported.

\section*{Conflict of interest}

The authors declare no potential conflict of interests.

\section*{Data Availability Statement}

The datasets analyzed during the current study are not publicly available due to patient confidentiality and privacy restrictions but are available from the corresponding author on reasonable request.

\section*{Ethics approval statement}

The study was conducted in accordance with the Declaration of Helsinki, and approved by the Ethics Committee of Medical Ethics Committee of Nanfang Hospital of Southern Medical University (NFEC-2024-328 and 1 July 2024).

\bibliographystyle{unsrtnat}
\bibliography{reference}

\end{document}